# Deep Learning the Physics of Transport Phenomena

Amir Barati Farimani,* Joseph Gomes,* and Vijay S. Pande†

*Department of Chemistry, Stanford University, Stanford, California 94305*

## Abstract

We have developed a new data-driven paradigm for the rapid inference, modeling and simulation of the physics of transport phenomena by deep learning. Using conditional generative adversarial networks (cGAN), we train models for the direct generation of solutions to steady state heat conduction and incompressible fluid flow purely on observation without knowledge of the underlying governing equations. Rather than using iterative numerical methods to approximate the solution of the constitutive equations, cGANs learn to directly generate the solutions to these phenomena, given arbitrary boundary conditions and domain, with high test accuracy (MAE<1%) and state-of-the-art computational performance. The cGAN framework can be used to learn causal models directly from experimental observations where the underlying physical model is complex or unknown.

Transport phenomena studies the exchange of energy, mass, momentum, and charge between systems,[1] encompassing fields as diverse as continuum mechanics and thermodynamics, and is used heavily throughout all engineering disciplines. Here, we show that modern deep learning models, such as generative adversarial networks, can be used for rapid simulation of transport phenomena without knowledge of the underlying constitutive equations, developing generative inference based models for steady state heat conduction and incompressible fluid flow problems with arbitrary geometric domains and boundary conditions. In contrast to conventional procedure, the deep learning models learn to generate realistic solutions in a data-driven approach and achieve state-of-the-art computational performance, while retaining high accuracy. Deep learning models for physical inference can be applied to any phenomena, given observed or simulated data, and can be used to learn and predict directly from experiments where the underlying causal model is complicated or unknown.

In recent years, there have been several advances in the fields of computer vision and natural language processing applications brought on by deep learning.[2–5] The convolutional neural network architecture, an example of a modern deep learning architecture, is a multilayer stack of optimizable convolution operations which compute non-linear input transformations. Each operation in the stack transforms its input in a manner that increases the selectivity and uniqueness of the output representation. The flexibility of deep neural networks allows models in principle to learn successively higher orders of features from raw data, making the application of deep learning models in physics highly attractive.

There are already several examples of note exploring the application of deep learning techniques within the physics and engineering communities,[6–10] including applications for accelerating fluid simulations in graphics generation[11,12] and shape optimization for drag reduction.[13] While previous approaches have shown deep learning models perform exceptionally well in classification and regression tasks, we show that deep learning models are also effective at generating realistic samples from arbitrary data distributions.

Generative adversarial networks[14,15] (GAN) are a class of deep learning model for learning and generating samples from a data distribution. The network is composed of two main functions: the generator $G(z)$, which maps a sample $z$ from a random uniform distribution to the desired distribution, and the discriminator $D(\hat{x})$, which determines if sample $\hat{x}$ belongs to the observed data distribution. The conditional GAN (cGAN) is an extension of GAN where both generator and discriminator receive conditioning variables, $c$, in addition to or in place of the sample $z$. Conditional GANs have been successfully used previously for style transfer, texture mapping, text to image translation, image to image translation.[16–18] We adapt the cGAN model architecture and training procedure, diagrammed in Figure 1a.

We construct models for steady state heat transfer and incompressible fluid flow that learn directly from observation. In the case of heat transfer, the observations are the temperature distribution given a set of constant temperature boundary conditions. In the case of incompressible fluid flow, the observations are the velocity fields and pressure distribution given a set of constant velocity boundary conditions. Details on the construction of train and test datasets are given in Methods (section B).

The evaluation of the trained heat transfer model on four representative samples from the test set is demonstrated in Figure 2. The results shows a very close prediction of the temperature field compared to ground truth. The difference between ground truth and generated solutions corresponds to average mean absolute deviations of 0.72% (see SI). The locations of level set change in temperature contours are nearly identical between the deep learning model and the ground truth. The solution of the heat equation on rectangular domains of different area and dimensions are shown in Figures 2a and 2b. The model successfully differentiates between inside and outside of the boundary, correctly infilling the solution. In the case of the annulus geometric domain (Figure 2c), the model detected both the curvilinear boundary condition and the fact that the domain is inside the annulus. For the triangular geometric domain example shown in Figure 2d, the temperature boundary condition is a trigonometric function of (x,y) along the boundary. The model successfully handles both domain and oscillatory boundary conditions.

The evaluation of the trained fluid flow model on four representative samples from the test set is shown in Figure 3. The input velocities are given in Figure 3 (left) and the directions are illustrated by arrows for each example. The combination of positive and negative $u$ and $v$ boundary conditions creates different flow patterns, such as rotational or jet, inside the cavity. The predicted (Figure 3, center) and ground truth (Figure 3, right) velocity fields $u$ and $v$ are shown by vector fields and the pressure field contours are overlaid. In all four test cases, the prediction is highly accurate for velocity and pressure fields. The flow patterns predicted by the model are nearly exact. One interesting point to note here is the accurate inference of the pressure field given only the velocity field boundary condition, highlighting the information transfer between velocity and pressure fields which occurs within our model.

For all problems in field-based physics, there is an underlying physical relationship, described by differential equations, between field values at discrete, adjacent nodes in space. Numerical solvers utilize this fact by discretization of the underlying constitutive equations in order to obtain the field solution self-consistently (for example, $\nabla^2$T=0 implies that T at each node is the average of four adjacent nodes). The cGAN model forms an estimate of the data distribution by treating the model underlying the observations as a Markov random field (MRF),[19–21] where each node is considered as a random variable. During training, the discriminator learns the local relationship between adjacent nodes in order to distinguish between real and generated samples, a conditional probability query of the data distribution. The generator learns to produce the most probable sample given known evidence (boundary conditions and domain), a *maximum a posteriori* query of the data distribution. The parameters of the generator are then optimized to maximize the probability that the discriminator assigns a real label to generated data.

In Figure 4, we demonstrate how the discriminator operates on an example selected from the heat transfer test set. The discriminator operates over patches of data in a given example, attempting to classify if each patch is real or fake. The discriminator correctly classifies most patches of the generated and ground truth solutions. Since the process of featurization and feature detection are unsupervised, any number of known or unknown physical phenomena can be learned from raw data by the discriminator. In the case of steady-state heat conduction, the cGAN discriminator learns to identify patches which locally satisfy the underlying partial differential equation (Methods, Equation 4), i.e. the temperature at each node is the average of its four neighboring nodes. In the case of incompressible fluid flow, the cGAN discriminator scans the data within a particular field (e.g. $u$ velocity field) as well as the coupling between fields (e.g. $u$ and $v$ velocity fields and pressure field). The cGAN discriminator learns to identify patches of multi-channel data that locally satisfy the coupled partial differential equations in Methods, Equation 5 (conservation of mass and momentum).

To demonstrate the accuracy of our framework, we compare the generated solutions from our cGAN model with the FD method solutions for the test set fraction of the heat transfer and fluid flow datasets. The generated solutions show an average per-sample mean absolute error of less than 1% (see SI). The evaluation time of the cGAN model is over an order of magnitude faster than the conventional finite difference method (see SI).

We have shown that conditional generative adversarial networks can be used to directly infer physics from observations with high fidelity and state-of-the-art computational performance. We demonstrated successful learning and prediction for steady state heat conduction and incompressible fluid flow, two popular and widely applied physical phenomena. We have argued that the cGAN model learns from observations by treating the underlying physical model as a Markov random field. Our framework is capable of generalizing the learned physics to unseen domain geometries and boundary conditions, making it amenable as a general physics prediction engine. In case of incompressible fluid flow, we have shown that multiple fields describing different physics learn from each other, making it possible to couple multiple physics simultaneously. These results indicate that cGAN models can learn and generalize any non-linear, multiphysics phenomena. As sensors and data acquisition devices and their connectivity continue to grow exponentially, we can expect our framework will be used in predicting complex multiphysics phenomena in a faster, less compute-intensive manner. We expect the cGAN method to be broadly applicable to a wide range of scientific and engineering fields.

## METHODS

### A. Generative Adversarial Networks

We adapt the conditional Generative Adversarial Network (cGAN), used previously for image-to-image translation.[22] cGANs are generative models that learn a mapping from observed data $c$ to output data $\hat{x}$: $G$: $c \rightarrow \hat{x}$. Here, $c$ is a representation of solution domain and boundary conditions and $\hat{x}$ is the observed solution. The generator $G(c)$ is optimized to produce outputs $\hat{x}$ that cannot be distinguished from training data by a discriminator, $D$. $D(c, \hat{x})$ is a scalar that represents the probability that $\hat{x}$ came from $p_{model}(\hat{x})$ (the data distribution) rather than the output of $G(c)$. The generator $G$ and discrimator $D$ models are convolutional neural networks, adapted from Ref. 23. The generator $G$ uses a "U-Net"-based network architecture[24] and the discriminator $D$ uses a convolutional "PatchGAN" classifier architecture.[25] The combined network architecture and training procedure are diagrammed in Figure 1a.

We train $D$ to maximize the probability of assigning the correct label to both training examples and samples from $G$. The cGAN objective function is expressed as:

$$\mathcal{L}_{cGAN}(G, D) = \mathbb{E}_{c,\hat{x}\sim p_{model}(c,\hat{x})}[logD(c, \hat{x})]+ \\ \mathbb{E}_{c,\hat{x}\sim p_{model}(c,\hat{x})}[log(1 - D(c, G(c)))]. \tag{1}$$

Here, $D$ and $G$ participate in a two-player minimax game with value function $\mathcal{L}_{cGAN}(G, D)$, where $G$ attempts to minimize this objective against an adversarial $D$ that tries to maximize it, $G^* = arg\ min_G\ max_D\ \mathcal{L}_{cGAN}$. In addition, we apply an L1 distance loss function to the generator:

$$\mathcal{L}_{L1}(G) = \mathbb{E}_{c,\hat{x}\sim p_{model}(c,\hat{x})}[\|\hat{x} - G(c)\|_1]. \tag{2}$$

The final objective is:

$$G^* = arg\ min_G max_D \mathcal{L}_{cGAN}(G, D) + \lambda\mathcal{L}_{L1}(G). \tag{3}$$

where the hyperparameter $\lambda$ is the L1 objective weight.

The discriminator $D$ (Figure 1c) is a convolutional neural network that operates on either ($c$, $\hat{x}$), the input and generated output, or ($c$, $\hat{x}'$), the input and ground truth solution, to produce the probability that a small patch of the discriminator input comes from the training data distribution. In this way, the discriminator treats each real or generated sample as a Markov random field (MRF), an undirected probabilistic graph which assumes statistical independence between nodes separated by more than a patch diameter.[19–21] This operation is performed convolutionally across the entire solution, averaging all responses of all distinct patches to provide the output of $D$. By training the discriminator to correctly distinguish between real and generated samples (Methods, Equation 2), we build a joint probabilistic model over the desired field values at discrete grid points. In the image modeling community, the treatment of images as MRFs has been previously explored, and is commonly used in models to determine texture or style loss.[26,27]

For generator $G$ (Figure 1b), we adopt an encoder-decoder network, which has been used successfully in image and text translation.[28–30] The input representing the geometric domain and boundary conditions is processed by the encoder network (Figure 1b, top), a series of convolutional operations that progressively downsample the grid until a reduced vector representation is reached (Figure 1d, top). The process is then reversed by the decoder network (Figure 1b, bottom), with a series of convolutional operations that progressively upsample the reduced representation, directly generating the inferred solution (Figure 1d, bottom). Due to the great deal of low-level information shared between the input and output grids, we share information directly between equivalent size encoder and decoder convolutional layers by the use of skip connections.[24] Generation proceeds by sampling the conditional probability density for the state of each grid point given the known states of its neighboring data points. By training a loss function that rewards the generator $G$ for successfully confusing the discriminator $D$, in addition to reproducing the ground truth solution for known observations, the generator is able to infer realistic solutions and generalize to situations (boundary conditions and domain shape/size) outside the scope of training data.

## B. Models

### 1. *Heat Transfer*

We consider steady-state heat conduction on an arbitrary two-dimensional domain with no heat generation. We wish to obtain a solution of the temperature field, $T(x, y)$. Here, the Laplace equation applies:

$$\frac{\partial^2 T}{\partial x^2} + \frac{\partial^2 T}{\partial y^2} = 0 \tag{4}$$

assuming constant thermal conductivity. The boundary conditions are expressed in the form $T(x, y) = f(x, y)$, where $f$ is a piece-wise continuous function.

A dataset containing 6,230 training samples was generated by numerical finite difference (FD) method, varying temperature boundary conditions, two-dimensional geometry (rectangle, disk, annulus, triangle), domain size, and domain position within a 64x64 grid. The training data consists of pairs of 1-channel, 64x64 grids; the first grid represents the solution domain and boundary conditions (Figure 2, left) and the second grid contains the solved temperature field (Figure 2, right). The portion of the grid outside of the solution domain is set a value of -1 in both input and solution. The dataset was split randomly into train/test sets following an 80/20 ratio. An additional dataset containing larger domain sizes (256x256 and 512x512) was generated in a similar manner for timing purposes.

*2. Fluid Mechanics*

We consider steady-state fluid flow in a square two-dimensional domain. We wish to obtain a solution of the velocity fields, $u(x, y)$ and $v(x, y)$, and pressure field, $p(x, y)$. Here, the non-linear Navier-Stokes equations apply:

$$\begin{aligned} \frac{\partial u}{\partial x} + \frac{\partial v}{\partial y} &= 0; \\ \rho(u\frac{\partial u}{\partial x} + v\frac{\partial u}{\partial y}) &= -\frac{\partial p}{\partial x} + \mu(\frac{\partial^2 u}{\partial x^2} + \frac{\partial^2 u}{\partial y^2}); \\ \rho(u\frac{\partial v}{\partial x} + v\frac{\partial v}{\partial y}) &= -\frac{\partial p}{\partial y} + \mu(\frac{\partial^2 v}{\partial x^2} + \frac{\partial^2 v}{\partial y^2}). \end{aligned} \tag{5}$$

We assume constant density $\rho$ and viscosity $\mu$. The boundary conditions are expressed in the form $u(x, y) = f(x, y)$ and $v(x, y) = g(x, y)$, where $f$ and $g$ are piece-wise continuous functions.

A dataset containing 4,850 training samples was generated by numerical FD method, varying velocity boundary conditions. The velocity boundary conditions (two u velocities in x-direction on the top and bottom lids, and two v velocities in y-direction for left and right lids) are are allowed to vary independently within a range of -2 and 2 m/s with the step size of 0.2 m/s. The training data consists of pairs of 3-channel, 64x64 grids; the first grid represents the solution domain and boundary conditions (Figure 3, left) and the second grid contains the solved velocity and pressure fields (Figure 3, right). The first two channels contain velocity, $u$ and $v$, and the third channel contains pressure, $p$. The pressure channel in the input data is initialized to 0. The dataset was split randomly into train/test sets following an 80/20 ratio.

## ACKNOWLEDGEMENTS

The Pande Group is broadly supported by grants from the NIH (R01 GM062868 and U19 AI109662) as well as gift funds and contributions from Folding@home donors. We acknowledge the generous support of Dr. Anders G. Frøseth for our work on machine learning.

## AUTHOR CONTRIBUTIONS STATEMENT

ABF and JG contributed equally to this work. VSP supervised the project.

## COMPETING FINANCIAL INTERESTS

VSP is a consultant and SAB member of Schrodinger, LLC and Globavir, sits on the Board of Directors of Apeel Inc, Freenome Inc, Omada Health, Patient Ping, Rigetti Computing, and is a General Partner at Andreessen Horowitz.

## SUPPORTING INFORMATION

The evaluation of train and test set statistical accuracy, computational performance, and diagram of the incompressible fluid flow cGAN CNN architecture are given in supporting information. The training data and model codes will be available upon request and as soon as paper is published.

---

[*] These authors contributed equally to this work

[†] Corresponding Author email: pande@stanford.edu

[1] R. B. Bird, W. E. Stewart, and E. N. Lightfoot, *Transport phenomena* (John Wiley & Sons, 2007).

[2] A. Krizhevsky, I. Sutskever, and G. E. Hinton, in *Advances in neural information processing systems* (2012) pp. 1097–1105.

[3] Y. Wu, M. Schuster, Z. Chen, Q. V. Le, M. Norouzi, W. Macherey, M. Krikun, Y. Cao, Q. Gao, K. Macherey, *et al.*, arXiv preprint arXiv:1609.08144 (2016).

[4] Y. LeCun, Y. Bengio, and G. Hinton, Nature **521**, 436 (2015).

[5] I. Goodfellow, Y. Bengio, and A. Courville, *Deep learning* (MIT press, 2016).

[6] J. Carrasquilla and R. G. Melko, Nature Physics **13**, 431 (2017).

[7] E. P. van Nieuwenburg, Y.-H. Liu, and S. D. Huber, Nature Physics **13**, 435 (2017).

[8] Q. Wei, R. G. Melko, and J. Z. Chen, Physical Review E **95**, 032504 (2017).

[9] K. Mills, M. Spanner, and I. Tamblyn, arXiv preprint arXiv:1702.01361 (2017).

[10] G. Carleo and M. Troyer, Science **355**, 602 (2017).

[11] S. Jeong, B. Solenthaler, M. Pollefeys, M. Gross, *et al.*, ACM Transactions on Graphics (TOG) **34**, 199 (2015).

[12] J. Tompson, K. Schlachter, P. Sprechmann, and K. Perlin, arXiv preprint arXiv:1607.03597 (2016).

[13] X. Guo, W. Li, and F. Iorio, in *Proceedings of the 22nd ACM SIGKDD International Conference on Knowledge Discovery and Data Mining* (ACM, 2016) pp. 481–490.

[14] I. Goodfellow, J. Pouget-Abadie, M. Mirza, B. Xu, D. Warde-Farley, S. Ozair, A. Courville, and Y. Bengio, in *Advances in neural information processing systems* (2014) pp. 2672–2680.

[15] I. Goodfellow, arXiv preprint arXiv:1701.00160 (2016).

[16] T. Kim, M. Cha, H. Kim, J. Lee, and J. Kim, arXiv preprint arXiv:1703.05192 (2017).

[17] J.-Y. Zhu, T. Park, P. Isola, and A. A. Efros, arXiv preprint arXiv:1703.10593 (2017).

[18] C. Ledig, L. Theis, F. Huszár, J. Caballero, A. Cunningham, A. Acosta, A. Aitken, A. Tejani, J. Totz, Z. Wang, *et al.*, arXiv preprint arXiv:1609.04802 (2016).

[19] Z. Wu, D. Lin, and X. Tang, in *European Conference on Computer Vision* (Springer, 2016) pp. 295–312.

[20] P. Acar and V. Sundararaghavan, Modelling and Simulation in Materials Science and Engineering **24**, 075005 (2016).

[21] Y. Lu, S.-C. Zhu, and Y. N. Wu, (2016).

[22] P. Isola, J.-Y. Zhu, T. Zhou, and A. A. Efros, arXiv preprint arXiv:1611.07004 (2016).

[23] A. Radford, L. Metz, and S. Chintala, arXiv preprint arXiv:1511.06434 (2015).

[24] O. Ronneberger, P. Fischer, and T. Brox, in *International Conference on Medical Image Computing and Computer-Assisted Intervention* (Springer, 2015) pp. 234–241.

[25] C. Li and M. Wand, in *European Conference on Computer Vision* (Springer, 2016) pp. 702–716.

[26] C. Li and M. Wand, in *Proceedings of the IEEE Conference on Computer Vision and Pattern Recognition* (2016) pp. 2479–2486.

[27] R. Yeh, C. Chen, T. Y. Lim, M. Hasegawa-Johnson, and M. N. Do, arXiv preprint arXiv:1607.07539 (2016).

[28] G. E. Hinton and R. R. Salakhutdinov, science **313**, 504 (2006).

[29] X. Wang and A. Gupta, in *European Conference on Computer Vision* (Springer, 2016) pp. 318–335.

[30] D. Yoo, N. Kim, S. Park, A. S. Paek, and I. S. Kweon, in *European Conference on Computer Vision* (Springer, 2016) pp. 517–532.

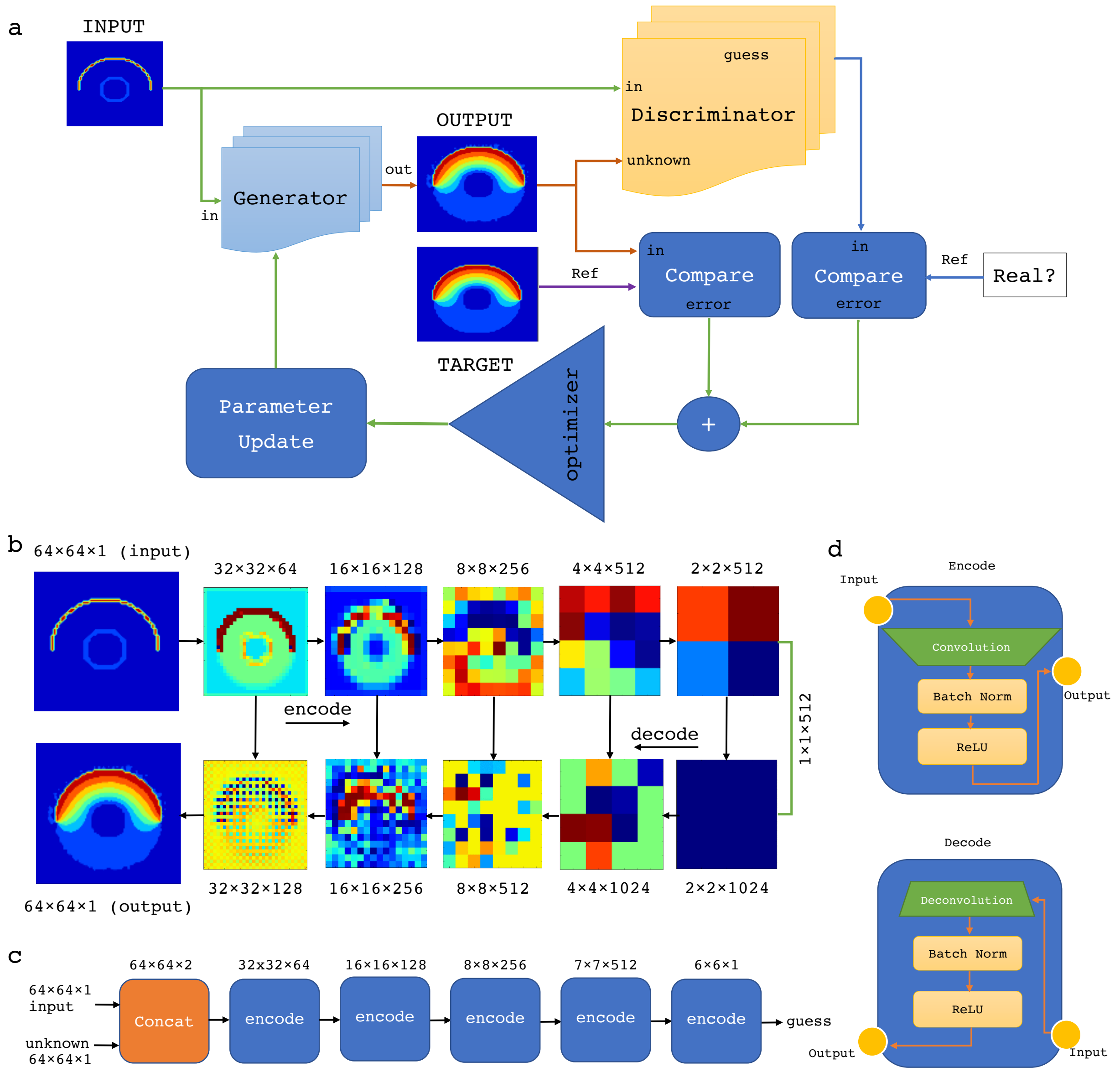

FIG. 1. **Conditional Generative Adversarial Network (cGAN) architecture. a.** The flowchart of cGAN and the connections between input (boundary conditions), output (field distribution), generator, discriminator, optimization operation and parameter update. **b.** The architecture of the convolutional neural network (CNN) used in the cGAN generator. A representative output for each layer is shown for both encode layers and decode layers. **c.** The architecture of the CNN used in the cGAN discriminator. The input (boundary conditions) and unknown (calculated or generated) solution are given to the discriminator which then guesses if the unknown solution is real or generated. **d.** The architecture of the encode and decode layers in cGAN.

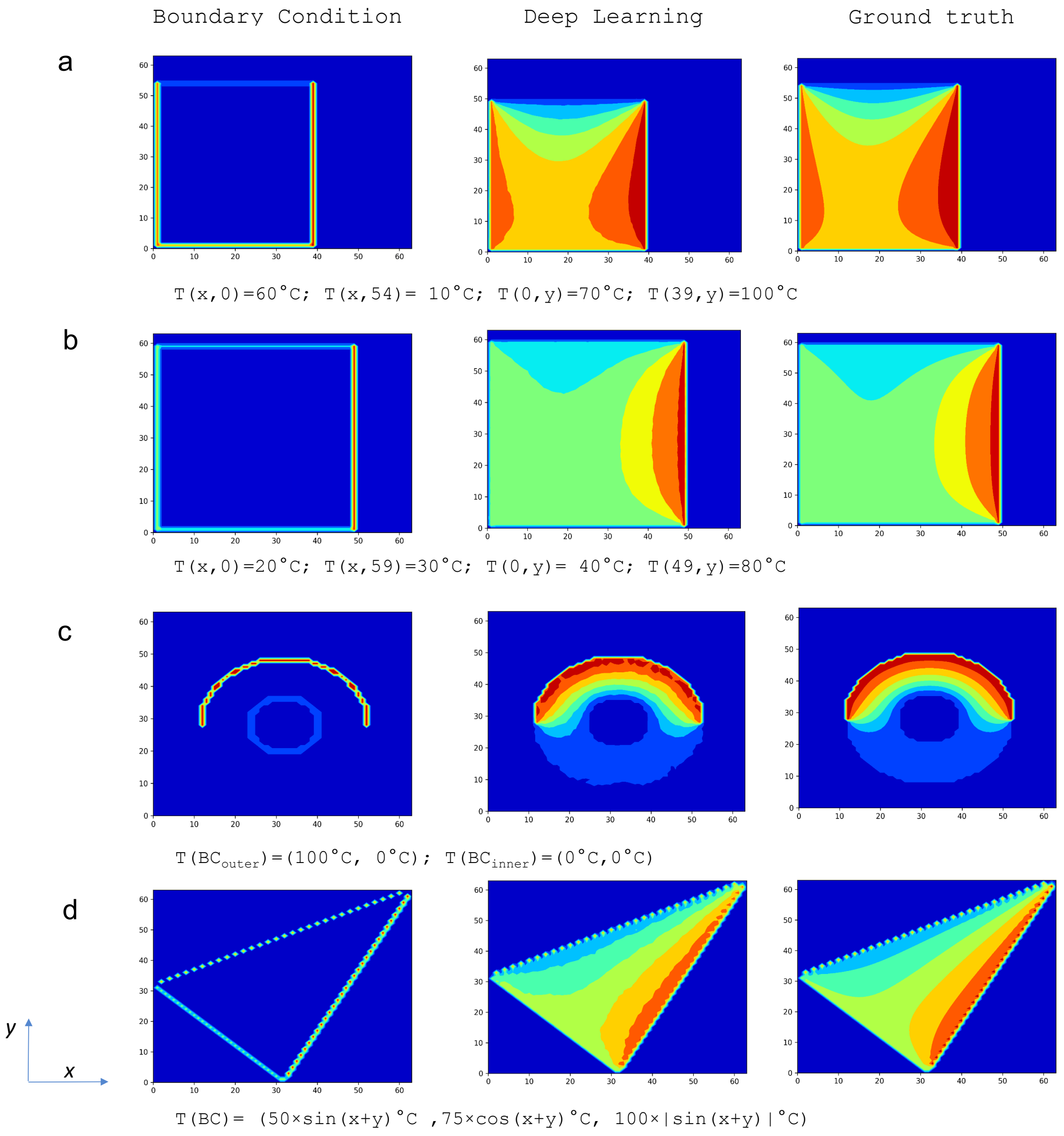

FIG. 2. **cGAN performance on heat transfer test set.** Steady-state heat conduction on a two-dimensional plane with constant temperature boundaries. The boundary condition input (left), the predicted solution (center), the ground truth solution of the temperature field (right) are shown for **a.** A rectangular domain with an offset from top and right sides. **b.** A rectangular domain with the smaller right boundary offset. **c.** An annular domain with different top and bottom temperatures. **d.** A triangular domain with oscillatory boundary conditions. Note that the only input to the deep learning model is the set of boundary conditions and no constitutive equation is given.

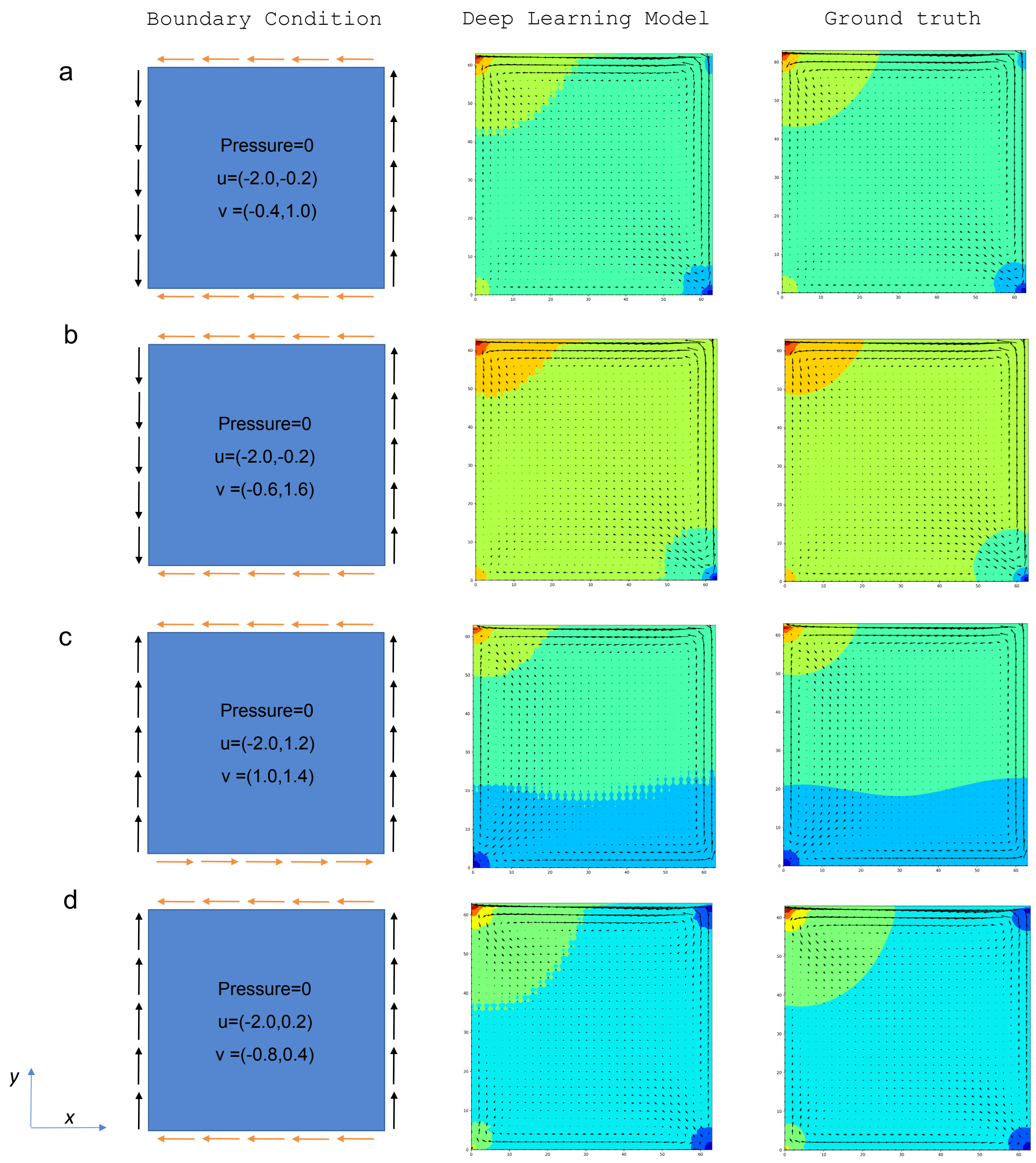

FIG. 3. **cGAN performance on fluid mechanics test set.** Fluid flow in a two-dimensional cavity with constant velocity boundaries. The boundary condition input (left), the predicted solution (center), and ground truth solution of the velocity (arrows) and pressure field (contour) (right) are shown for test samples **a-d**.

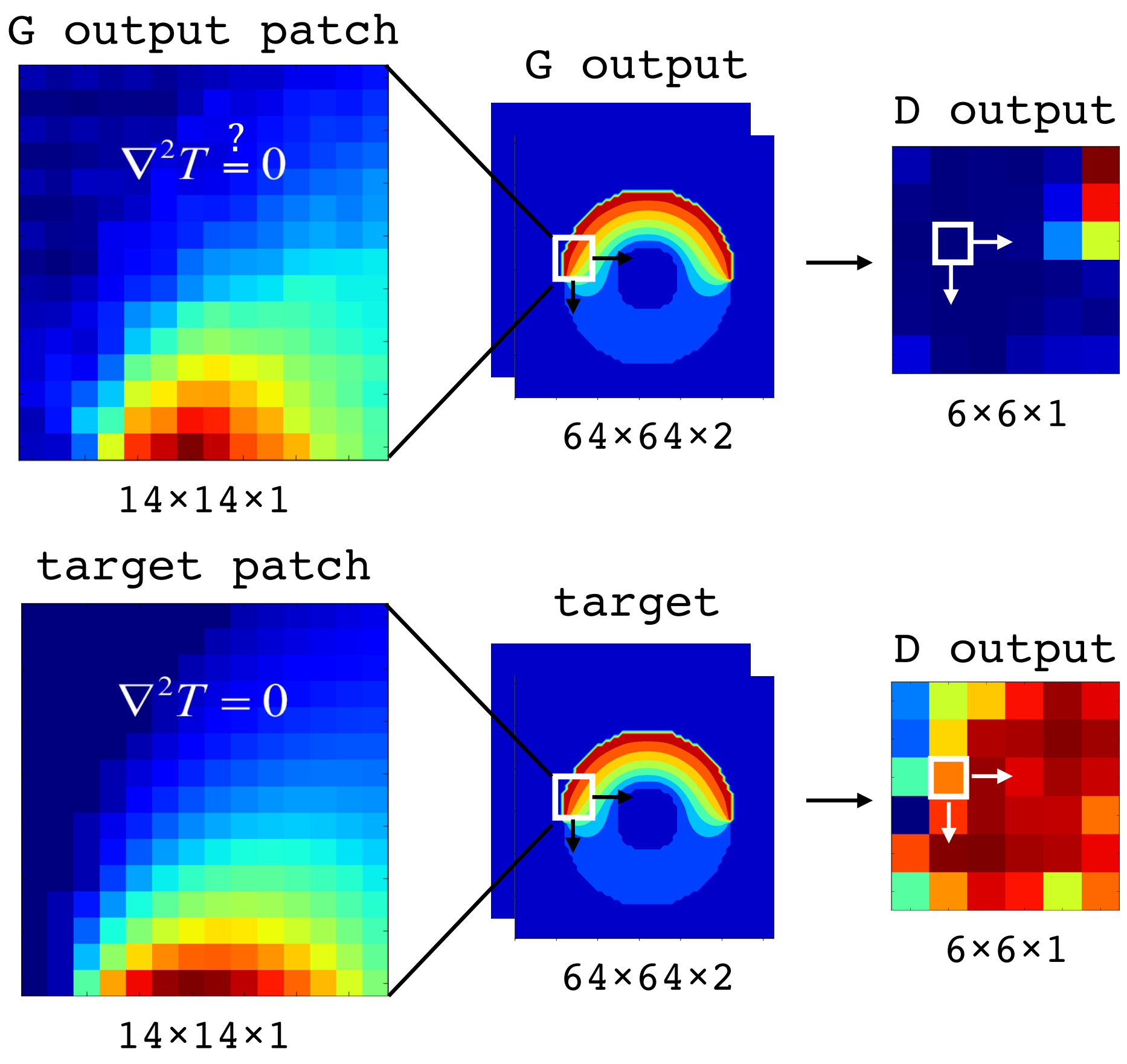

FIG. 4. **Learning physics by discrimination.** PatchGAN discriminator scans the data ($G$ output or target) and builds a joint probabilistic model of the data ($D$ output). Each element of $D$ output represents the probability that a 14x14 patch of $G$ output or target (shown on the left) comes from the true data distribution (red = high probability, blue = low probability). In the heat transfer example, the joint probability distribution is inferred as "each data point is the average of its four neighbors", which satisfies the PDE describing the temperature field. We show that the discriminator successfully classifies real and generated samples.

# Supporting Information for:

# Deep Learning the Physics of Transport Phenomena

Amir Barati Farimani,* Joseph Gomes,* and Vijay S. Pande†

*Department of Chemistry, Stanford University, Stanford, California 94305*

## I. EVALUATION OF HEAT TRANSFER AND NAVIER-STOKES MODELS ON TRAINING DATA.

The evaluation of the trained heat transfer model on four representative samples from the training set is demonstrated in Figure S1. The results shows good reproduction of the temperature field compared to ground truth. The evaluation of the trained fluid flow model on three representative samples from the training set is shown in Figure S2. The input velocities are given in Figure S2 (left) and the flow vector directions are illustrated by arrows for each example. The ground truth (Figure S2, center) and predicted (Figure S2, right) velocity fields $u$ and $v$ are shown by vector fields and the pressure field contours are overlaid.

Additionally, we assess the statistical accuracy of cGAN model predictions for all the samples used in **training set** of the heat transfer and fluid flow datasets (Figure S3). The maximum temperature error is 2.0°C across a test set with a temperature range of 100°C. The average per-sample root-mean squared deviation (RMSD) is 0.454°C, which less than 0.5% relative error. The maximum error in the predicted velocity fields 0.06 m/s across a test set with a velocity range of 4 m/s. The average per-sample RMSD for the $u$ and $v$ fields are 0.0128 m/s and 0.0129 m/s, a relative error of 0.32%. The maximum error in the predicted pressure field is 0.13 Pa across a test set with a pressure range of 40.0 Pa. The average per-sample RMSD For the pressure field, the average RMSD is 0.08786 Pa (0.22% relative error).

## II. EVALUATION OF HEAT TRANSFER AND NAVIER-STOKES MODELS ON TEST DATA.

To demonstrate the generalizability of our framework, we compare the generated solutions from our cGAN model with the FD method solutions for the test set fraction of the heat transfer and fluid flow datasets (Figure S4a-c). The maximum temperature error is 2.2°C across a test set with a temperature range of 100°C. The average per-sample mean absolute error (MAE) is 0.7226°C, which less than 1% relative error. The maximum error in the predicted velocity fields is 0.055 m/s across a test set with a velocity range of 4 m/s. The average per-sample RMSD for the $u$ and $v$ fields are 0.01809 m/s and 0.01355 m/s, a relative error of 0.35-0.45%. The maximum error in the predicted pressure field is 0.13 Pa across a test set with a pressure range of 40.0 Pa. For the pressure field, the average RMSD is 0.08782 Pa, a relative error of 0.22%. The generated solutions also satisfy the continuity, x-momentum and y-momentum equations (Main Text, Equation 5) with the the averaged convergence residuals of 0.00218 (continuity), 0.00264 (x-momentum), and 0.00308 (y-momentum), demonstrating the validity of generated fluid field solutions. We conclude that the deviations between the ground truth and the deep learning solution are very small. The difference between train and test error is almost negligible, suggesting good model generalizability.

## III. COMPUTATIONAL PERFORMANCE

We test the computational cost of the cGAN method directly with that of the FD method for obtaining solutions to the two-dimensional heat transfer problem. A comparison between the generation time for the cGAN model with the iterative solution time of the FD method, for a range of residual tolerance, across three grid sizes ($64^2$, $256^2$, $512^2$) are presented in Figure S4d. Timings were performed on a single core of an Intel Core i5-7400 CPU @ 3.00GHz with 16 Gb RAM. Details on the FD method algorithm and code are available in Ref.[1,2]. The particular code[3] was chosen as the fastest publically available serial finite difference code. Across all grid sizes, the cGAN method shows over an order of magnitude faster evaluation time compared with the FD method. The cGAN model possesses both a lower polynomial scaling and scaling prefactor; thus, the cGAN model outperforms the FD model across all possible grid sizes and the computational performance gain improves as the grid size is increased.

The accuracy of the current generative models are determined solely by model architecture and training procedure. Thus, by modification of the training procedure (model hyperparameters and training data), it is possible to improve model accuracy while maintaining constant evaluation time. Additionally, models can potentially be improved through optimization of the model architecture, which we leave to future work. We note that the generated solutions can be further refined by standard FD method techniques if greater accuracy is required. This procedure essentially uses the generative model as a initial guess initializer that can bypass many steps of iterative refinement and may improve self-consistent convergence for difficult systems.

## IV. NAVIER-STOKES MODEL NETWORK ARCHITECTURE.

The architecture of the CNN used in the cGAN generator for the 2D heat transfer model are shown in Figure 1 of the main manuscript. The architecture of the CNN used in the cGAN generator for the 2D fluid flow model is shown in Figure S5. The main modification of the fluid flow CNN is the addition of three input and output channels, due to the necessity of solving pressure and velocity fields simultaneously. In contrast to the heat transfer model, which only requires one input and out channel for $T$, we use three channels to describe the two velocity fields ($u$ and $v$) and the pressure field $p$. We note that for solving other multiphysics problems, one channel per field can be used.

---

* These authors contributed equally to this work

† Corresponding Author email: pande@stanford.edu

[1] S. Patankar, *Numerical heat transfer and fluid flow* (CRC press, 1980).

[2] M. Quinn, *Parallel Programming in C with MPI and OpenMP*, McGraw-Hill Higher Education (McGraw-Hill Higher Education, 2004).

[3] "2d steady state heat equation in a rectangle," `http://people.sc.fsu.edu/~jburkardt/cpp_src/heated_plate/heated_plate.html`, accessed: 2017-08-23.

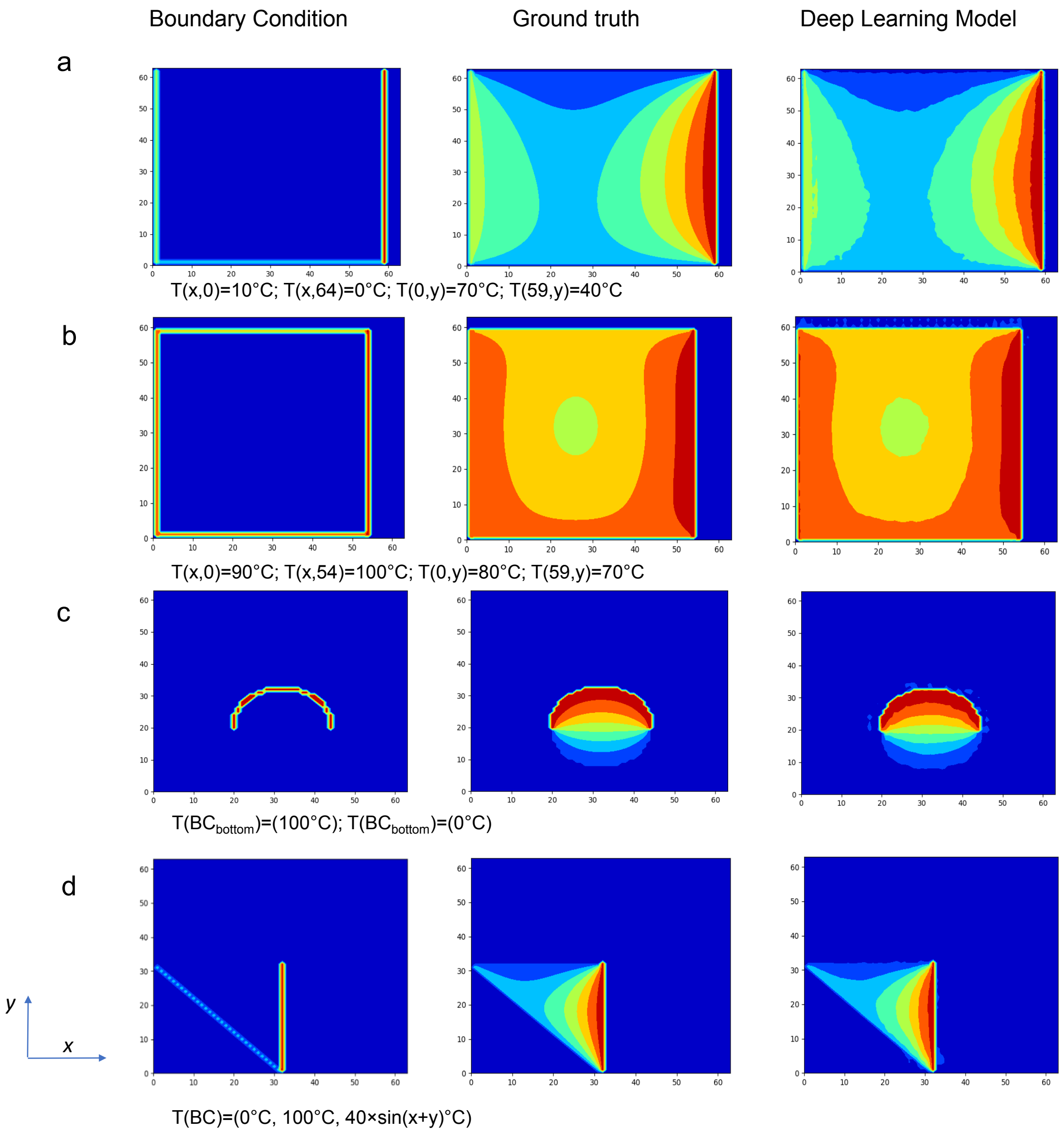

Figure S 1. **cGAN performance on heat transfer training set.** Steady-state heat conduction on a two-dimensional domain with constant temperature boundaries. The boundary condition input (left), the ground truth solution of the temperature field (middle), and the predicted Deep Learning (cGAN) results (right) are shown for **a.** A rectangular domain with an offset from top and right sides, **b.** A rectangular domain with the larger right boundary offset, **c.** A disc domain with different top and bottom temperatures, and **d.** A triangular domain with oscillatory boundary conditions.

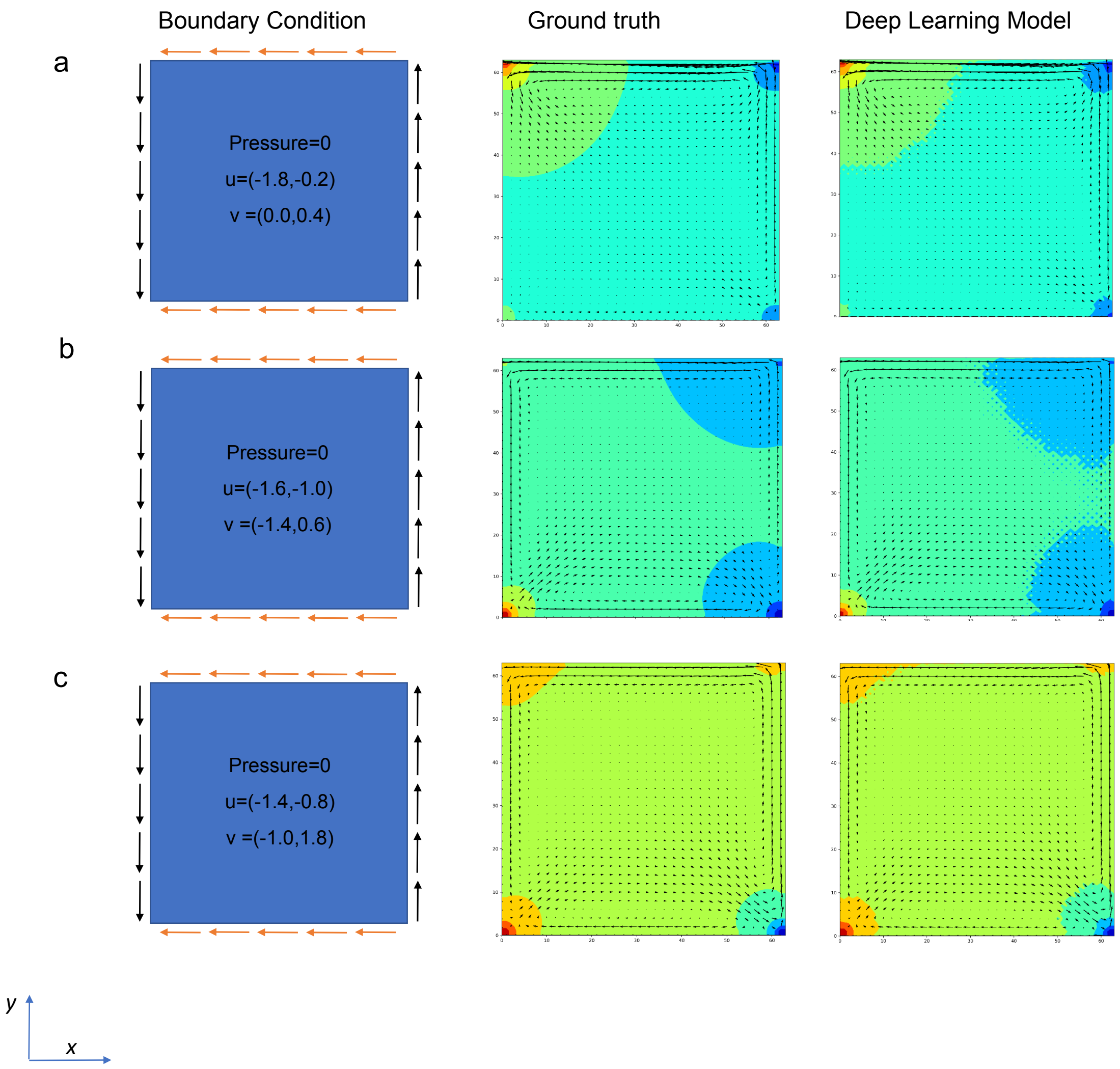

Figure S 2. **cGAN performance on fluid mechanics training set.** Fluid flow in a two-dimensional cavity with constant velocity boundaries. The boundary condition input (left), ground truth solution of the velocity (arrows) and pressure field (contour) (middle), and predicted Deep Learning (cGAN) results (right) are shown for training samples **a-c**.

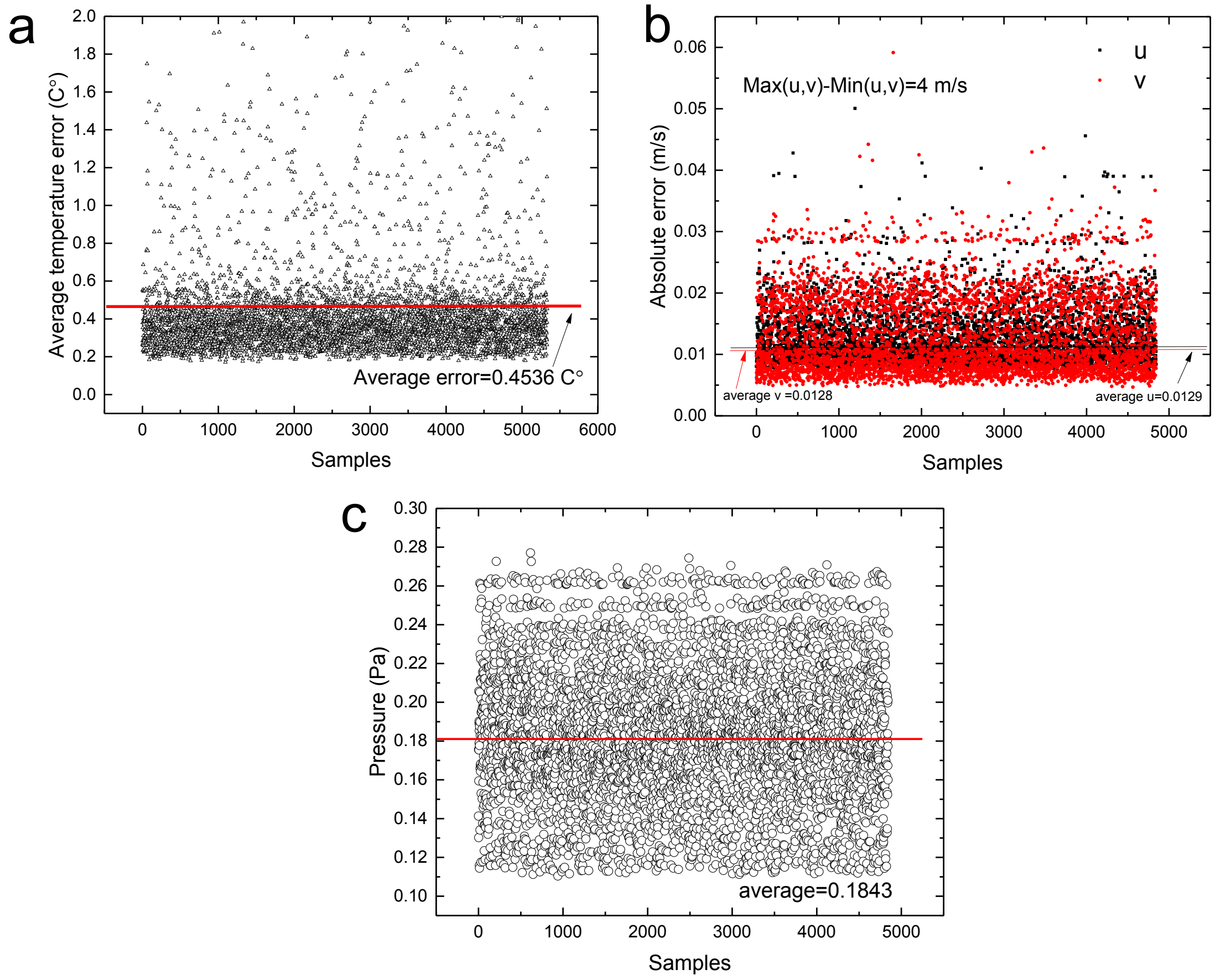

Figure S 3. **Statistical accuracy of cGAN model evaluated on the training set.** Per-sample root-mean squared deviation over **a.** heat transfer training set, **b.** fluid mechanics training set (velocity field), **c.** fluid mechanics training set (pressure field).

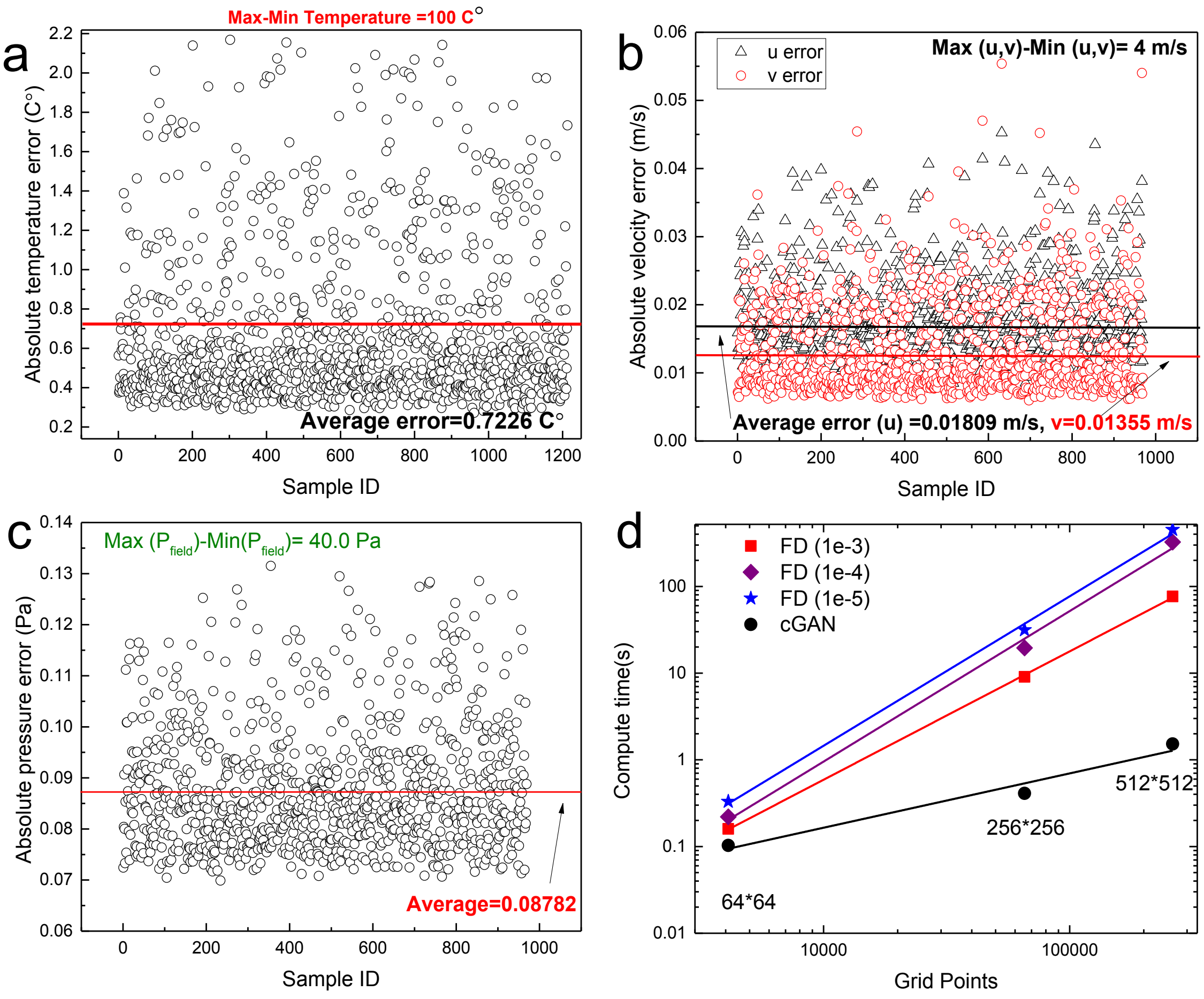

Figure S 4. **Statistical accuracy and computational performance.** Per-sample root-mean squared deviation over **a.** heat transfer test set, **b.** fluid mechanics test set (velocity field), **c.** fluid mechanics test set (pressure field). **d.** Representative timings of heat transfer test set solution. Average time (s) for solution at varying grid size. A comparison between finite difference methods and generative method.

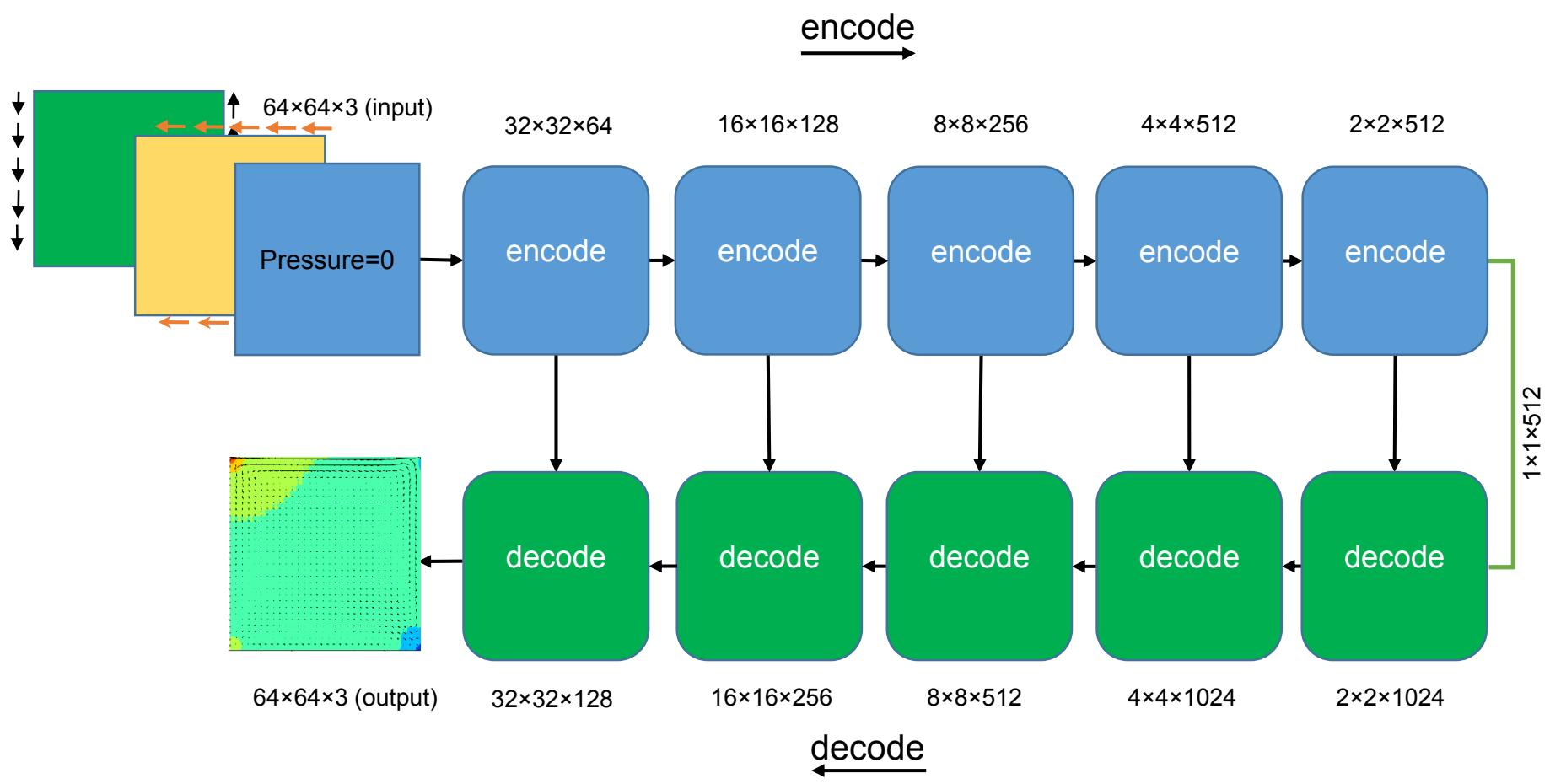

Figure S 5. **cGAN CNN architecture for fluid flow model.** The architecture of the CNN used in the cGAN generator for the fluid flow model.